# Automated Heterogeneous Low-Bit Quantization of Multi-Model Deep Learning Inference Pipeline

Jayeeta Mondal
TCS Research
Kolkata, India
jayeeta.m@tcs.com

Swarnava Dey
TCS Research
Kolkata, India
swarnava.dey@tcs.com

Arijit Mukherjee
TCS Research
Kolkata, India
mukherjee.arijit@tcs.com

## Abstract

*Multiple Deep Neural Networks (DNNs) integrated into single Deep Learning (DL) inference pipelines* e.g. *Multi-Task Learning (MTL) or Ensemble Learning (EL), etc., albeit very accurate, pose challenges for edge deployment. In these systems, models vary in their quantization tolerance and resource demands, requiring meticulous tuning for accuracy-latency balance. This paper introduces an automated heterogeneous quantization approach for DL inference pipelines with multiple DNNs.*

## 1. Introduction

As the pursuit of more precise Deep Learning (DL) solutions advances, larger and increasingly intricate Deep Neural Networks (DNNs) are being introduced. This poses challenges when attempting to deploy them on edge or compact hardware platforms that have limitations in terms of storage, memory, and energy consumption. Moreover, there exists a considerable demand for on-device DL inference aimed at enhancing user experiences and bolstering data security. This is evident in applications such as human detection and facial recognition within surveillance cameras, prediction and recognition systems embedded in smartphones, object recognition utilizing LiDAR point-cloud data in autonomous vehicles, etc. Additionally, to attain the utmost precision in fully automated Artificial Intelligence (AI) systems, it's common practice to integrate multiple DNNs into a unified DL inference pipeline, as seen in approaches like Multi-Task Learning (MTL) or Ensemble Learning (EL) solutions [2, 9]. Certain models within these inference pipelines can tolerate more aggressive quantization, while others might necessitate more resource-intensive operations. Achieving computational efficiency in such Deep Learning (DL) systems involves the intricate task of manually investigating various levels of quantization [7, 3]. This process aims to strike the right balance between accuracy and latency, but it's a laborious undertaking with extended time-to-market (TTM) implications. There exists some earlier works on multi-level quantization within a DNN [12, 1, 5, 4]. Coelho *et al.* [5] applied multi-level quantization to individual layers in a deep neural network (DNN) handling real-time data from high-speed proton-proton collisions. Yet, examining every layer of each DNN isn't essential, particularly in multi-task or ensemble setups where their contributions to the overall deep learning inference pipeline can differ, potentially leading to sub-optimal results. In this paper, we introduce an automated quantization approach. It examines potential quantization levels for all models in a multi-model deep learning pipeline and recommends the optimal quantization level for each based on the desired overall system accuracy and latency.

## 2. Methodology

In this section of the paper, we formulate our problem mathematically and represent the proposed methodology. Consider: 1) $M$: A set of $N$ models in a multi-model DL inference pipeline; 2) $Q$: A set of different levels of low-bit quantization possible to perform in a full precision DNN; 3) $A$: A set of $N$ accuracy thresholds for each model in set $M$; 4) $L$: A set of $N$ latency thresholds for each model in set $M$. The optimization problem aims to find the optimal quantization levels that balance accuracy and latency for all models in the pipeline. Mathematically, the problem can be formulated as,

$$\max \sum_{i=1}^{N} \sum_{q \in Q} b_{i,q} \cdot A_i; \min \sum_{i=1}^{N} \sum_{q \in Q} (1 - b_{i,q}) \cdot L_i; \quad (1)$$

Equation 1 is subjected to the following conditions: for each model $i \in M$ enforce that either the accuracy or latency constraint is satisfied; ensure that each model $i$ is quantized at only one level; and put binary constraint on the decision variables.

$$\sum_{q \in Q} b_{i,q} \cdot (A_i - L_i) \geq 0; \sum_{q \in Q} b_{i,q} = 1; b_{i,q} \in \{0, 1\} \quad (2)$$

Algorithm 1 seeks optimal quantization levels for N models in a DL inference pipeline, aiming to balance maximizing accuracy and minimizing latency. It uses binary decision variables to determine the quantization state, then iterates to calculate the total accuracy and latency. Here, the constraints enforce accuracy-latency trade-offs and single-level quantization. The multi-objective optimization problem is solved to obtain the best quantization decisions.

---

**Algorithm 1** Multi-Objective Optimization for determining quantization levels of each model in a multi-model DL inference pipeline.

**Input:** $N, Q, A, L$
**Output: quantization levels for all $N$ models in set $M$**
Initialize binary decision variables $b[i][q]$
Initialize variables for accuracy sum *accuracy_sum* and latency sum *latency_sum*
**for** $i$ in $N$ **do**
    **for** $q$ in $Q$ **do**
        *accuracy_sum* += $b[i][q] \cdot A[i]$
        *latency_sum* += $(1 - b[i][q]) \cdot L[i]$
    **end**
**end**
Define constraints:
**for** $i$ in $N$ **do**
    Append inequality constraint:
    $\sum_{q \in Q} b[i][q] \cdot (A[i] - L[i]) \geq 0$
**end**
**for** $i$ in $N$ **do**
    Append equality constraint:
    $\sum_{q \in Q} b[i][q] = 1$
**end**
Initialize initial guess: $b[i][q] = 0$
Solve using multi-objective optimization solver
Extract optimal quantization decisions:
*optimal_quantization* ← $b[i][q]$

---

## 3. Experimental Results and Conclusion

In this section of the paper we use our methodology to perform heterogeneous multi-level quantization of the 3 DNNs used in Google's MediaPipe [11] hand gesture recognizer DL inference pipeline. The 3 DNNs include: 1) *MHD*: an object detection convolutional neural network (CNN) for hand detection, 2) *MKD*: a CNN based key-point detection model, and 3) *MKC*: a mutli-layered perceptron (MLP) for key-point classification. In our experiments we fix the quatization levels set $Q$ = (fp-16, int-8, int-4, int-2, bin-1). Here, we quantize full-precision floating-point 32-bit models to *fp-16*: floating-point 16-bit; *int-8, int-4, int-2*: integer 8-bit, 4-bit and 2-bit respectively; and *bin-1*: binarized 1-bit models. 1-bit quantization is extreme quantization that allows replacing MAC operations in CNNs with XNOR [6, 8, 10]. We measure the accuracy of each individual DNN in MediaPipe on the data collected using Raspberry Pi Camera Rev 1.3 in Raspberry Pi 3 Model B and find the inference speed on it. The plot presented in Figure 1 illustrates the relationship between inference speed (min-max normalized) and accuracy for the models within the MediaPipe hand gesture recognition deep learning pipeline, quantized at different levels. By applying the methodology detailed in the preceding section, we ascertain the optimal quantization levels for each model, as highlighted in Table 1.

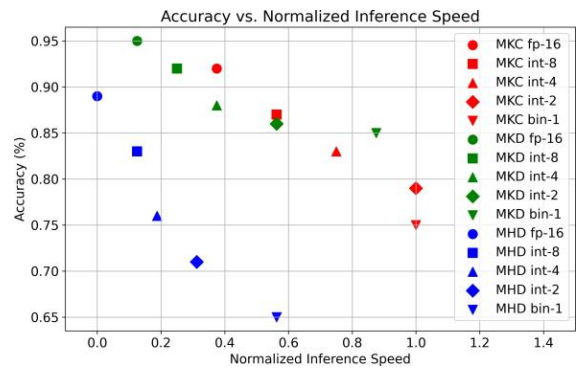

Figure 1. Accuracy Vs Inference speed plot for heterogenous quantization of various components of Mediapipe [11] hand gesture recognition DL inference pipeline.

| Model | Parameter Precision | Accuracy | size (KB) | latency (ms) |
|---|---|---|---|---|
| MHD | fp-16 | 0.88 | 2500 | 250 |
| MKD | bin-1 | 0.85 | 171 | 20 |
| MKC | int-8 | 0.87 | 500 | 130 |

Table 1. Performance evaluation of heterogeneously quantized DL pipeline of MediaPipe [11] using methodology in Section 2 measured in Rpi3.

Employing the proposed automated quantization technique, we achieved a 12x boost in inference speed for the MediaPipe hand gesture recognizer, accompanied by a minor 4% accuracy reduction. Consequently, this approach proves advantageous for deploying multi-model DL systems on resource-constrained platforms.